# A Hybrid Strategy for Chat Transcript Summarization

Pratik K. Biswas

*Abstract*—Text summarization is the process of condensing a piece of text to fewer sentences, while still preserving its content. Chat transcript, in this context, is a textual copy of a digital or online conversation between a customer (caller) and agent(s). This paper presents an indigenously (locally) developed hybrid method that first combines extractive and abstractive summarization techniques in compressing ill-punctuated or un-punctuated chat transcripts to produce more readable punctuated summaries and then optimizes the overall quality of summarization through reinforcement learning. Extensive testing, evaluations, comparisons, and validation have demonstrated the efficacy of this approach for large-scale deployment of chat transcript summarization, in the absence of manually generated reference (annotated) summaries.

## I. INTRODUCTION

Automatic document summarization aims to compress a textual document to a shorter, more informative format while keeping key information of the original text. Hence, automatic text summarization has become a very desirable tool in today's information age. Numerous approaches have been developed for automatic text summarization that can be broadly classified into two groups: extractive summarization and abstractive summarization. Extractive summarization extracts important sentences from the original text and reproduces them verbatim in the summary, while abstractive summarization generates new sentences. Hybrid Summarization attempts to combine these two approaches in some form.

Chat transcript is a record of the digital or online conversation between a caller (customer) and receiver(s) (customer representatives/agents). In this paper, however, we shall be confining ourselves to *online chats* of a *phone company*. These are not exchanges of *tweets* (postings) or *text messages* of restricted lengths, but unrestricted textual dialogs about problems and resolutions. Automatic summarization of chat transcripts (as with call transcripts), in our domain, pose certain unique challenges, as follows: *1) they are not continuous texts but include conversation between customers and agents, 2) they are either very short or very long, and can include a large number of sentences that are irrelevant and even meaningless, 3) they include several ill-formed, grammatically incorrect sentences, 4) they are either un-punctuated or are improperly punctuated, 5) there is a dearth of a large collection of human-crafted reference (annotated) summaries and 6) existing open-source summarization tools don't perform too well with chat transcripts.* Hence the need for *a new domain-specific chat summarizer* that can meet the needs of a *phone company*.

How do we achieve large-scale (industrial grade) summarization of domain specific chat transcripts (dialogs) in the absence of annotated (reference) summaries? How do we find the most effective LM/LLM-based, fine-tuned abstractive summarizer for this purpose? In this paper, we address these two questions by introducing a novel, hybrid summarization technique that combines *channel separation* (separation into customer and agent transcripts), *topic modeling, sentence selection*, and *punctuation restoration* based extractive summarization and *transfer learning* based abstractive summarization with *contextual multi-armed bandits* based *reinforcement learning* to generate, validate and optimize properly punctuated, fixed-length and readable customer and agent summaries from the original chat transcripts, that can adequately summarize customer concerns and agent resolutions.

## II. RELATED WORK

Related research can be broadly grouped into three categories: 1) extractive summarization, 2) abstractive summarization and 3) hybrid summarization.

Radev et al. [1] defined summary as "a text that is produced from one or more texts, that conveys important information in the original text(s), and that is no longer than half of the original text(s) and usually, significantly less than that." Automatic text summarization gained attraction as early as the 1950s. Different methods and extensive surveys of automatic text summarization have been provided in [2]-[7].

Luhn et al. [8] introduced a method to extract salient sentences from the text using features such as word and phrase frequency. They proposed to weigh the sentences of a document as a function of high frequency words, ignoring very high frequency common words. Gong et al. [9] and Wang et al. [10] described how multiple documents could be summarized using topic models. Xu et al. [11] constructed a neural model for single-document summarization based jointly on extraction and syntactic compression. Miller [12] used Bidirectional Encoder Representations from Transformers (BERT) for summarization of lecture notes. Liu [13] described BERTSUM, a simple variant of BERT, for extractive summarization. Liu et al. [14] showcased how BERT could be generally applied in text summarization and proposed a general framework for both extractive and abstractive models. Tuggener et al. [15] provided an extensive overview of existing dialog summarization data sets and mappings from data sets to linguistic models. Feigenblat et al. [16] introduced TWEETSUM, a large-scale

P. K. Biswas is with Artificial Intelligence and Data Science, Verizon Communications, Basking Ridge, NJ, USA 08854 (e-mail: Pratik.Biswas@verizonwireless.com).

database of customer support dialogs with extractive and abstractive summaries along with an unsupervised extractive summarization method specific to these dialogs.

Lin et al. [17] surveyed the state-of-the-art while Khan et al. [18] reviewed the various methods in abstractive summarization. Nallapati et al. [19], Paulus et al. [20], See et al. [21] and Liu et al. [22] employed recurrent neural networks, deep reinforcement learning, pointer-generator, and generative adversarial networks for abstractive summarization. Rafael et al. [23] showed the effectiveness of transfer learning in abstractive text summarization through a unified framework named as Text-to-Text-Transfer-Transformer (T5). Lewis et al. [24] developed BART, a denoising autoencoder for pretraining sequence-to-sequence models that is particularly effective when fine-tuned for text generation, e.g., abstractive summarization, translation. Beltazi et al., [25] presented Longformer, useful for long document summarization. Zhang et al. [26] proposed PEGASUS, where during pre-training important sentences were removed from an input document and the model was tasked at recovering them. Gliwa et al. [27] introduced the SAMSum Corpus, a new dataset with abstractive dialog summaries and showed that model-generated summaries of dialogs performed better than the model-generated summaries of news; while Khalman et al. [28] collected ForumSum, a diverse and high-quality conversation summarization dataset to show that models trained on ForumSum transferred better to new domains compared to SAMSum. Fabbri et al. [29] crowdsourced four new datasets from news comments, discussion forums, community question answering forums, email threads and benchmarked state-of-the-art abstractive summarization models on their datasets. Zhong et al. [30] unveiled DialogLM, a denoising, pre-trained, neural encoder-decoder model for long dialog understanding and abstractive summarization.

Chen and Bansal [31] followed a hybrid "extract-then-rewrite" architecture, with policy-based RL to fuse the extractor agent and the abstractor network. Bae et al. [32] rewrote sentences from a document and then paraphrased the selected ones to generate a summary. Su et al. [33] combined the two summarization methods and a text segmentation model to generate variable-length, fluent summaries.

The proposed hybrid strategy is a significant extension of our earlier work on extractive summarization [43]. It consists of multiple phases. The first phase is an adaptation, from the published extractive strategy, for domain-specific chat transcript summarization, while the remaining *phases* are completely new additions to this manuscript.

## III. MAJOR CONTRIBUTIONS

Our main contributions and advantages can be summarized as follows:
- We combine topic modeling, embedding-based sentence selection and transformer-based punctuation restoration for unsupervised extractive summarization.
- We fine-tune powerful, transformer-based language models, on indigenously (locally) extracted summaries, for abstractive summarization of chat transcripts through transfer learning and identify the fine-tuned abstractive summarizer that is most effective for our transcripts.
- We apply contextual multi-armed-bandits-based reinforcement learning to validate the results of our summarization and optimize the overall quality of the chat summaries.

The summaries can be used both as *historical records* and *reminder messages* of prior chats.

## IV. PRELIMINARIES – CONCEPTS AND TERMINOLOGIES

In this section, we clarify key concepts and terminologies and explain certain technologies, which provide the foundation for our work.

### A. Automatic Summarization

The automatic summarization of text is a well-defined task in the field of *Natural Language processing (NLP)*. *Automatic text summarization* attempts to convert a larger document into a shorter form preserving its information content and overall meaning. A good summary should reflect the diverse topics of the document while keeping redundancy to a minimum. Next, we look at the two different approaches to automatic text summarization, namely extractive and abstractive.

*Extractive summarization* methods condense long documents into shorter versions by directly extracting sentences from the relevant sections in the original text. In contrast, *abstractive summarization* methods attempt to convey the main ideas from the original text through new sentences, and not by copying verbatim the most important sentences from the same text. In other words, they interpret, examine, and analyze the original text using advanced natural language techniques to get a better understanding of the content and then describe it through shorter and more focused text, comprising of new sentences. Purely extractive summaries often give better results than automatic abstractive summaries, because abstractive summarization methods deal with semantic representation, inference and natural language generation, which are relatively harder than sentence extraction. Many abstractive summarization techniques, particularly the ones using deep learning, often depend upon the extracted summaries as the labeled data for building the training samples on which to train to generate new text.

### B. Topic Modeling

*Topic modeling* is an unsupervised machine learning method for finding a group of words (i.e., topic), from a collection of documents, that best represents the information in the collection. Many techniques have been used to obtain probabilistic topic models. *Latent dirichlet allocation (LDA)* is a popular topic modeling technique that represents documents as a random mixture of latent topics, where each topic is a probability distribution of words [34]. The *hierarchical dirichlet process (HDP)* is another topic modeling technique that extends LDA. It is a nonparametric Bayesian approach that uses a mixed-membership model for unsupervised analysis of grouped data. Unlike LDA (its finite counterpart), HDP infers the number of topics from the input data. *Latent semantic analysis (LSA)*, also known as *latent semantic indexing (LSI)*, is an *unsupervised* method for extracting a representation of text semantics from observed words [35]. It attempts to bring out latent relationships among a collection of documents on to a lower-dimensional space. LSA is based on the principle that words closer in meaning occur in similar pieces of text (the distributional hypothesis). It has been used for multi-document summarization.

*C. Transformers*

*Transformers* in NLP provide general-purpose, novel, neural network architectures for *Natural Language Understanding (NLU)* and *Natural Language Generation (NLG)* with over 32+ pre-trained language models. They were first introduced in [36]. Transformers are *Seq2Seq deep learning* models that transform sequential inputs to sequential outputs, while handling long-range dependencies with ease. They are based solely on *attention* mechanisms, dispensing entirely with *recurrence* and *convolutions* of the earlier deep learning architectures. Transformers do not require that the sequential data be processed in order, which allows for much more parallelization than *Recurrent Neural Networks (RNNs)* and therefore reduced training times [36]. Since their introduction, transformers have become the model of choice for tackling many problems in NLP, replacing older recurrent neural network models such as the *Long Short-term Memory (LSTM)*. Transformer models can train on much larger datasets than before, as they can support more parallelization during training. This has resulted in the development of pre-trained systems such as *Bidirectional Encoder Representations from Transformers (BERT)*. BERT is a bidirectional transformer pre-trained, using a combination of *masked language* modeling objective and next sentence prediction, on a large corpus comprising the Toronto Book Corpus and Wikipedia, by jointly conditioning on both left and right contexts in all layers. Consequently, a pre-trained BERT model can be fine-tuned with just one additional output layer to create state-of-the-art models for a wide range of NLP tasks [37]. *GPT-2* [38] *and XLNet* [39] are two other recently pre-trained NLP models. GPT-2 was trained on a 40 GB dataset called WebText and has approximately 1 billion parameters. It was trained to predict the next word. XLNet is an improved version of BERT that implements *permutation language* modeling in its architecture and randomly predicts the next tokens.

Transformers often employ a 12-layered encoder-decoder architecture consisting of one stack of 6 encoding and another stack of 6 decoding layers. The encoders are all identical in structure, each composed of two sub-layers, namely *self-attention,* and *feed-forward neural network*. The decoder additionally has an *attention* layer, in between the two, that helps it to focus on relevant parts of the input sentence (similar to what attention does in *Seq2Seq models*). When we pass a sentence into a transformer, it is embedded and passed into a stack of encoders, which processes the input iteratively one layer after another. The output from the final encoder is then similarly processed through each decoder layer (block) in the decoder stack, eventually generating the output.

*D. Embeddings*

*Embeddings* are mathematical functions that map "entities" to a latent space with complex and meaningful dimensions. Words or sentences or paragraphs can be mapped into a shared latent space such that the meaning of the word/sentence/paragraph can be represented geometrically. Machine learning approaches towards NLP require words to be expressed in vector form. Word embeddings is a feature engineering technique in which words are mapped into a vector of real numbers in a pre-defined vector space. It is a learned representation for text, where words with the same meaning have similar representations. The idea of using a densely distributed representation for each word is a key to this approach. *Word2Vec, GloVe*, etc. provide pre-trained word embedding models in a type of *transfer learning*. Embedding techniques initially focused on *words* but later shifted attention to other types of textual content, such as *n-grams*, *sentences,* and *documents*. The *Universal Sentence Encoder (USE)* encodes text into high dimensional vectors used for text classification, semantic similarity, clustering, and other natural language tasks [40]. The model is trained and optimized for sentences, phrases, or short paragraphs, from a variety of data sources with the aim of dynamically accommodating a wide variety of natural language understanding tasks. The model maps variable length input English text into an output of a 512-dimensional vector.

*E. Contextual Multi-armed Bandits*

*Multi-Arm bandit (MAB)* [41] is a classic *reinforcement learning* problem, in which a player faces a slot machine or bandit with $k$ arms, each with an unknown payout. Pulling any one of the arms fetches the player either a reward or a penalty and he/she can pull only a limited number of times. The player's objective is to pull the arms one-by-one in sequence to maximize the total reward collected over time. At the beginning of the experiment, when odds and payouts are unknown, the player must determine which arm to pull, in which order and how many times. The solution involves exploration/exploitation tradeoff: the player should balance between trying different arms to learn more about the expected payout of every arm and taking advantage of the best arm that he/she knows.

The *contextual multi-armed bandit (CMAB)* algorithm [42] is an extension of the multi-armed bandit approach where we factor in the customer's environment, or context, when choosing a bandit's arm. The context affects how a reward is associated with each arm, so as contexts change, the model should learn to adapt its bandit choice, that is, the selection of its arm. These algorithms can adapt and learn from the environment automatically, without human intervention.

*Bandit testing* applying MAB/CMAB can be thought of as an alternative to A/B testing. Unlike A/B testing where traffic is evenly split between two variations, MAB/CMAB allows the user to dynamically allocate traffic to better performing variations while allocating less and less traffic to underperforming ones.

## V. HYBRID SUMMARIZATION OF CHAT TRANSCRIPTS

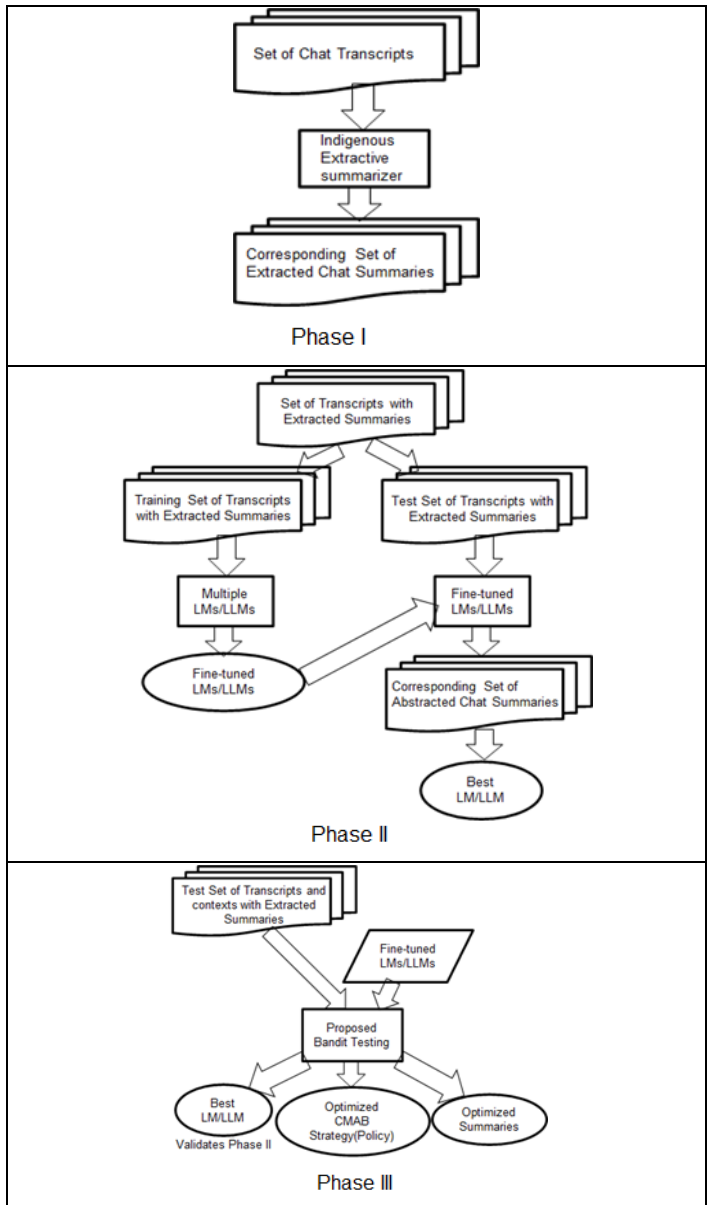

**Fig. 1.** Hybrid Summarization of Chat Transcripts (3 Phases)

We propose a hybrid strategy that extracts, fine-tunes, abstracts, validates, and optimizes. The main objective of this strategy is to provide a hybrid summarization framework that accomplishes the following: i) extracts *good quality* summaries of chat transcripts to create a large enough training sample; ii) uses this sample to fine-tune pre-trained, transformer model based abstractive summarizers to generate new summaries of unseen transcripts on the fly through *transfer learning*, where the resultant summaries are at least *as good as the extractive summaries* (on which the abstractive summarizers have been fine-tuned), with the tacit expectation that the pre-training encoded in the abstractive approach would make the summaries more fluent, coherent and even help reduce some grammatical errors found in the original transcripts; iii) supports reinforcement learning, allowing us to summarize with multiple fine-tuned abstractive summarizers and helps not only in finding the best performing summarizer for a large number of transcripts but also the right variant for a given transcript, improving the overall quality of summarization in the process. So, the strategy involves 3 sequential phases. Phase I uses an extractive summarizer, Phase II uses abstractive ones, and Phase III optimizes the overall quality of the summaries. The strategy is driven by the fact that it will be used in a production environment, which requires the summarization of a very large number of chat transcripts. but where there is a paucity of *manually generated reference (annotated)* summaries from which the abstractive summarizers can learn in Phase II and with which we can compare our results. **Fig. 1** shows the 3 Phases of the proposed hybrid summarization strategy.

### A. Phase I – Sample Generation through Extractive Summarization

Phase I generates a large pool of chat summaries, through *extractive summarization*, which can be reused for fine-tuning abstractive summarizers. This extractive summarization technique uniquely integrates *channel (speaker) separation*, *topic modeling*, and *similarity-based sentence selection* with *punctuation restoration* through a 10-step sequential procedure, as shown in **Fig. 2**. It is based on an adaptation from our earlier work on extractive summarization of *call transcripts* [43].

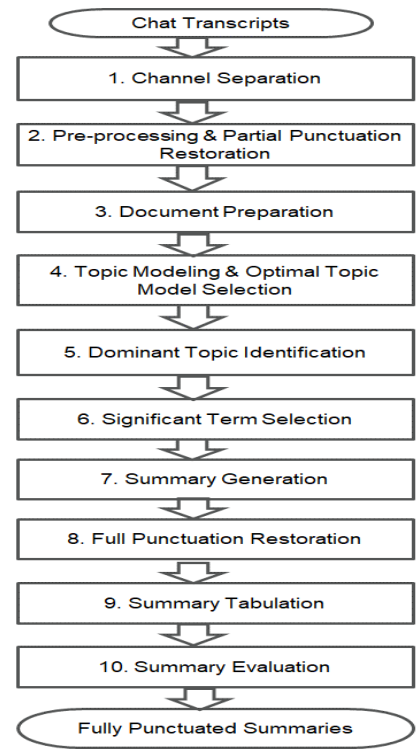

**Fig. 2.** Phase I – Extractive Summarization.

The *punctuation restored summaries* are the outputs from this procedure. The procedure is highly parameterized. The full list of parameters to the proposed procedure includes: *Topic Model Type* (default: "None/False"), *Number of Topics* (default: 5), *Number of Dominant Topics* (default: 1), *Batch Size for Punctuation Restoration* (default: 512), *Term Extraction Method* (default: "global"), *Desired Summary Length* (default: 5), *Summary Table Name* (default: "summary_results"), *Word Similarity Threshold* (default: 0.5), *Uniqueness Threshold for Sentence Similarity* (default: 0.5). Next, we describe the key steps of this indigenous (our) extractive summarization process.

*1) Channel Separation*

Chat transcripts include conversations/dialogs between a customer and one or more agents and so the resultant summaries can often get mixed up. So, separation of a transcript into customer and agent transcripts, based on channel or speaker identifier, can make each summary more coherent. **Algorithm 1** is a very basic algorithm for splitting a full chat transcript, consisting of multiple utterances with their corresponding channel or speaker identifiers, into corresponding customer and agent transcripts based on the type of the identifier.

If the channel identifiers, associated with the transcripts, are not available (as they are sometimes absent) to clearly identify the speakers then we can use a *pre-trained BERT Transformer* model [37] with a *linear classifier* from *PyTorch nn* package as an additional layer, on top of BERT's 12 layers, to classify each dialog of the transcript into one of the 2 classes, i.e., *customer* and *agent* and then combine each type of dialogs to create customer and agent transcripts.

| **Algorithm 1** Channel Separation |
|---|
| $u$: list of all utterances in a conversation |
| $s$: list of the corresponding speaker identifiers for the utterances |
| results: set of separated customer and agent transcripts |
| set: finds a set from a list (removing duplicates) |
| len: computes the length of a string |
| range: returns a sequence of numbers, starting from 0 and stopping at a specified number |
| cat: concatenates strings |
| **Inputs**: $u, s$ <br> **Output**: results |
| count ← 0 <br> results ← {} <br> **for** sp ∈ set($s$) **do** <br>    **for** $i$ ∈ range(len($s$)) **do** <br>       **if** $s_i$ is sp **then** <br>          **if** count is 0 **then** <br>             trans ← $u_i$ <br>             count ← count + 1 <br>          **else** <br>             trans ← cat(cat(trans,"."), $u_i$) <br>    results ← results ∪ {trans} <br>    count ← 0 <br> **return** results |

*2) Document Preparation*

A *document* is a list of *keywords* extracted from each transcript and is used as input to the topic model. For document preparation, we build a custom NLP preprocessing pipeline comprising of tokenization; punctuation, extended stop-words, and small words (length ≤ 4) removal; regular expression matching; lowercasing; contraction mapping; bigrams and trigrams creation; lemmatization; parts of speech tagging and allowable tag selection. This has been implemented by combining modules available from four *Python* packages, namely, *re*, *spaCy*, *NLTK* and *gensim*.

*3) Topic Modeling and Optimal Topic Model Selection*

If the *topic model type* is specified at the invocation of the procedure, then we create multiple *topic models* of the desired type, for both customer and agent, using the *documents*, *corpus* and *vocabulary* from the corresponding chat transcripts, by varying the hyper-parameter (e.g., *topic number*) values within the pre-defined ranges by the pre-defined steps; compute their coherence scores and identify the topic models and associated hyper-parameter values that produce the best scores. Otherwise, by default, we perform the above-mentioned activity for all 3 different topic model types, namely, *LDA*, *LSI* and *HDP*, in parallel, and identify the topic models and associated hyper-parameter values that produce the best scores amongst topic models of all three types. **Algorithm 2** is a simplified algorithm for the selection of the optimal topic model from a set of either chat (full) or customer or agent chat (separated) transcripts.

| **Algorithm 2** Topic Model Optimization and Selection |
|---|
| $T$: input set of chat/customer/agent transcripts |
| $d$: documents from $T$ |
| $v$: vocabulary from $T$ |
| $c$: corpus from $T$ |
| tmg: function for generating Topic Models (TM) |
| $n$: topic number (hyper-parameter of topic model) |
| $t$: TopicModelType (summarizer parameter) |
| $N$: NumberOfTopics (summarizer parameter) |
| ccs: computes coherence score |
| otm: optimal Topic Model |
| **Inputs**: $d, v, c, t, N$ <br> **Output**: otm |
| **if** TopicModelType **then** <br>    otm = argmax$_{tm \in TM}$(ccs(tmg($v, c, d, t, n$)): $n = N, .., 50$) <br> **else** <br>    otm = argmax$_{t \in \{LDA,LSI,HDP\}}$(ccs(argmax$_{tm \in TM}$(ccs <br>                     (tmg($v, c, d, t, n$)): $n = N, .., 50$))) <br> **return** otm |

We have exclusively used the Python based *gensim* package for this step.

*4) Punctuation Restoration*

The punctuation restoration algorithm is used in steps 2 and 8 of the aforesaid procedure. In step 2, we preprocess transcripts (customer & agent) to remove existing punctuations and then restore punctuations *partially*, i.e., restore only *periods* as delimiters, so that sentences can be separated in each transcript and subsequently extracted to generate the summaries; while in step 8, we remove *existing periods* from each pair of customer and agent summaries, restore *full punctuation* and post-process to make them more readable. **Algorithm 3** describes the steps needed for punctuation restoration.

We have used the *BertForMaskedLM* class of the *PyTorch BERT* model (*bert-base-uncased*) [37, 46] for punctuation restoration and added an additional *linear* layer (*PyTorch nn module*) above the 12 BERT layers. The output of original BERT layers is a vector with the size of all vocabulary. The additional linear layer takes this as input and gives as output one of four classes, i.e., "O" (Other), "Comma", "Period" and "Question" for each encoded word. We retrained this modified BERT model using *TED transcripts*, consisting of two million words. Different variations of punctuation restoration with BERT model have been presented earlier but the retraining with the proposed architecture is a unique approach for punctuation restoration.

| **Algorithm 3** Punctuation Restoration |
|---|
| txt: text (chat/customer/agent transcript/summary) |
| segment_size: size of each segment |
| $conv_1$: converts tokens to ids |
| cs: creates segments with token_ids |
| bp: predicts using retrained BERT model |
| $conv_2$: converts predicted punctuation classes to characters |
| t_restored: restored and punctuated text |
| **Inputs**: $t$, segment_size, bp |
| **Output**: t_restored |
| tokens ← tokenize(txt) |
| token_ids ← $conv_1$(tokens) |
| smts ← cs(token_ids, segment_size) |
| preds ← bp(smts) |
| words_with_puncs ← $conv_2$(token_ids, preds) |
| t_restored ← merge_tokens(words_with_puncs) |
| **return** t_restored |

We found that the BERT model for punctuation restoration gave 30% more accurate results than the LSTM based model. We have implemented the punctuation restoration algorithm using *BERT Transformer*, *BertPunc* and *nn* packages, available from *PyTorch*.

*5) Summary Generation through Sentence Selection*

This process combines steps 5 through 7 of the main procedure, i.e., dominant topic identification, significant term selection and summary generation. First, for every transcript we get the specified number of *dominant topic(s)* from the selected topic model (for chat, customer, or agent) along with the associated keywords from the corresponding document (preprocessed transcript). Second, we do a *similarity analysis* between the dominant topics' aggregated list of *global* keywords/terms and the corresponding document's list of *local* transcript-based words to extract the most significant inter-related *terms* for each transcript and construct a string/document with them. Lastly, we generate fixed-length customer and agent summaries for every transcript by first identifying the most unique sentences in each of customer and agent transcripts based on *similarity analysis* among all sentences of the corresponding transcript using sentence *embeddings* and then selecting a fixed number (user-specified) of most relevant sentences from the condensed transcript through *sentence-based similarity analysis*, using *embeddings*, between every sentence of that transcript and the string/document constructed at the previous step. **Algorithm 4** explains this summary generation.

For *term-based similarity analysis*, we calculated *cosine similarity* between *GloVe* encoded word vectors (300 dimensions) using *spaCy*'s *en_vectors_web_lg*; while for *sentence-based similarity analysis and summary generation*, we used the *Universal Sentence Encoder (USE)* [40] from *tensorflow-hub*, and *Python* based *pandas* and *numpy* packages.

*6) Summarization Evaluation*

We determine the effectiveness of the summarizer by measuring both the *goodness* (*quality*) of summarization and the *correctness* (*accuracy*) of the punctuation restoration reflecting the readability of the summaries.

For the *quality* of the information content of the generated summaries, we use the metrics *BLEU* (Bilingual Evaluation Understudy) and *ROUGE* (Recall-Oriented Understudy for Gisting Evaluation) scores as a measurement of their *goodness*. We have compared the customer and agent summaries against the corresponding transcripts (or manually generated summaries if available) and computed their individual *BLEU and ROUGE-l* scores using the *Python* packages *NLTK* (*nltk.translate.bleu_score*) and *ROUGE* (*rouge*).

For the *correctness/accuracy* of the punctuation restoration, we can use the *accuracy_score* function from python's *sklearn.metrics* package to measure the *punctuation-restoration-accuracy* [43] as the number of matches of punctuation symbols (periods) between the original/extracted text (transcript/summary) and the punctuated text (transcript/summary), expressed as a percentage. We have computed the customer and agent *BLEU*, *ROUGE,* and *punctuation-restoration-accuracy* scores for every pair of customer and agent summaries, by iterating through all transcripts, and calculating their *averages* from their respective individual scores.

| **Algorithm 4** Summary Generation |
|---|
| $t$: transcript (chat/customer/agent) |
| otm: optimal Topic Model for the set of $t$ |
| $c$: corpus from $t$ |
| $N$: NumberOfDominantTopics (summarizer parameter) |
| $W$: WordSimilarityThreshold (summarizer parameter) |
| $U$: UniqenessThreshold (summarizer parameter) |
| $l$: desired summary length |
| preproc: preprocesses a transcript to return a list of words (document) |
| dom_topic: identifies $N$ dominant topics from a transcript and returns a list of keywords associated with them |
| flatten: flattens a list of keyword pairs |
| get_unique_sent: gets unique sentences from a transcript |
| $similarity_w$: does word-based similarity analysis |
| $similarity_s$: does sentence-based similarity analysis |
| extract_sum: extracts $l$ sentences from a transcript |
| sum: generated summary |
| **Inputs**: $t$, otm, $c$, $N$, $W$, $l$ |
| **Output**: sum |
| doc ← preproc($t$) |
| dom_kwds ← dom_topic(otm, c, doc, $N$) |
| sig_terms ← flatten($\{(k_1, k_2) : (k_1, k_2) \in$ doc × dom_kwds, $similarity_w((k_1, k_2) \geq W)\}$) |
| t_reduced ← get_unique_sent($t$, U)) |
| sum ← extract_sum($similarity_s$(t_reduced, sig_terms), $l$) |
| **return** sum |

## B. Phase II – Abstractive Summarization through Transfer Learning

Phase II involves fine-tuning pre-trained *Language/Large Language Model* (*LM/LLM*) driven, transformer based *abstractive summarizers* on the extractive summaries, obtained from Phase I, and then using these *fine-tuned summarizers* in generating summaries from unseen chat transcripts, via *transfer learning*, to find the most effective summarizer for our transcripts for potential production deployment.

We fine-tune *five* pre-trained LM/LLM based transformers, i.e., **T5** (*T5-small*) [50], **PEGASUS** (*pegasus-large*) [51], **BART** (*bart-large-xsum*) [52], **Longformer2Roberta** (*longformer2roberta-cnn_dailymail-fp16*) [53] and **DialogLED** (*DialogLED-large-5120*) [54] on the summaries extracted in Phase I, for abstractive summarization of chat transcripts.

**T5** [23, 49] and **PEGASUS** [26, 51], both encoder-decoder models, the former pre-trained on "Colossal and Cleaned version of Common Crawl" (C4) and the latter on C4 and "HugeNews" with "extracted gap sentences", can encode strings at most 512 and 1024 tokens respectively. **BART** (XSum) [24, 52] is a denoising transformer encoder-decoder (seq2seq) model with a bidirectional (BERT-like) encoder and an autoregressive (GPT-like) decoder, fine-tuned on the Extreme Summarization (XSum) dataset. It encodes upto 512 tokens. **Longformer2Roberta** [25, 53] is an encoder-decoder model, where the encoder is an *allenai/longformer-base-4096* model, and the decoder is a *roberta-base* model. It can handle upto 4096 tokens. It was fine-tuned on the CNN/DailyMail dataset, common for text summarization. **DialogLED** [30, 54] is a pre-trained model for long dialog understanding and summarization. It builds on the Longformer-Encoder-Decoder architecture and uses window-based denoising as the pre-training task on a large amount of long dialog data, encoding up to 5120 tokens. We have selected Google/**T5** as it is state-of-the-art, multi-purpose, and lightweight; Google/**PEGASUS** and Facebook/**BART** as they are commonly used for dialog summarization; **Longformer2Roberta** as many chat transcripts are long documents and **DialogLED** as it is designed to improve dialog summarization for long dialogs.

We apply the 5 abstractive summarizers, *with and without* (**ablation study**) fine-tuning, on unseen customer and agent transcripts, generate abstractive summaries and compare their performances, using the extractive summaries from the previous phase (Phase I). on *quality* (using metric scores, i.e., *BLEU and ROUGE* scores) and *fine-tuning times*. We identify the abstractive summarizer that performs the best for our chat transcripts and estimate the extent of quality improvement due to fine-tuning for each selected abstractive summarizer.

## C. Phase III – Validation and Optimization through Reinforcement Learning

Phase III deploys the 5 fine-tuned abstractive summarizers, validates the findings from Phase II by identifying the most competent abstractive summarizer amongst the 5 fine-tuned transformer models, through *Reinforcement Learning*, while attempting to improve the overall quality of chat transcript summarization for large-scale production deployment.

We attempt to *optimize* or *nearly optimize* the overall quality of the abstractive summaries for large number of chat transcripts through judicious use of all 5 fine-tuned abstractive summarizers, namely **T5**, **PEGASUS**, **BART**, **Longformer2Roberta** and **DialogLED**. For this purpose, we apply multiple policies (strategies) of *contextual multi-armed bandits* (CMAB) [42], a type of *reinforcement learning*, and also identify the best summarizer to validate the results from Phase II. CMAB adaptively *explores* and *exploits* the five fine-tuned summarizers on the chat transcripts to optimize on the metric (*BLEU* or *ROUGE*) scores of the summaries. During exploration phase, the algorithm randomly picks a summarizer to learn about its performance; while during the exploitation phase, the algorithm takes advantage of the summarizer that it knows to be the best. The algorithm acts by examining the context of each transcript, choosing one of the five summarizers to summarize the transcript and getting a reward (i.e., the corresponding metric score). The algorithm attempts to maximize the *average metric score* over all the transcripts while personalizing the selection of a summarizer for each transcript. The full algorithm, i.e., **Algorithm 5** is presented below.

---

**Algorithm 5** CMAB (Reinforcement Learning) for Chat Transcript Summarization

---

$K$: number of Summarizers (5)
$a$: Action (Selection of T5, PEGASUS, BART, Longformer2Roberta or DialogLED)
$r$: reward (BLEU or ROUGE score) in [0, 1]
$t$: Transcript
$c(t)$: context (Transcript attributes)
$e$: extracted summary
$T$: set of Transcripts
$E$: set of Extracted Summaries
$Q(a)$: average reward for Summarizer(action) $a$
$N(a)$: number of times summarizer $a$ has been used
$N$: number of times different summarizers been used
AMS: Average Metric Score
CMABp: CMAB Policy (algorithm)
zip: zips two sets/lists

---

**Inputs**: *S*ummarizers (T5 & LR), CMABp, $K, T, E$
**Output**: Best $a$ (Summarizer), AMS (Average Metric Score)

---

$N \leftarrow 0$, AMS $\leftarrow 0$
**for** $a = 1$ to $K$ **do**
    $Q(a) \leftarrow 0$, $N(a) \leftarrow 0$
**for** $t, e \in$ zip$(T, E)$ **do**
    $a \leftarrow$ CMAB$_p(c(t))$
    $r \leftarrow$ get_metric_score$(a(t), e))$
    $N(a) \leftarrow N(a) + 1$
    $Q(a) \leftarrow Q(a) + 1/N(a) * (r - Q(a))$
    $N \leftarrow N + 1$
    AMS $\leftarrow$ AMS $+ 1/N * (r -$ AMS$)$
**return** argmax$_a Q(a)$, AMS

## VI. PERFORMANCE EVALUATION

Effectiveness (quality of summaries), efficiency (summarization & fine-tuning time), flexibility, performance comparison with/among open-source, off-the-shelf summarizers, validation, and manual comparisons are some of the considerations that helped us evaluate the performance of our strategy for chat transcript summarization. The evaluation of results from all phases has been both automatic and manual.

## A. Experimental Setup

We set up a Spark cluster, consisting of a driver node and dynamically allocated, multiple executor nodes for data collection, preprocessing and summarization. We retrained/fine-tuned and tested the transformer models on NVIDIA Tesla V100-SMX2-32GB and A100-SMX4-40GB GPU based nodes. The driver node used anywhere between 1 to 4 GPUs. CUDA versions from 10.0 to 12.2 and NVIDIA CUDA Deep Neural Network (cuDNN) versions from 7.6 to 9.2.1 accelerated model training during different stages of the development of our hybrid strategy.

## B. Datasets

In Phase I, we have evaluated our extractive summarizer on 160,000 chat transcripts, covering a wide range of issues/topics including *billing, refunds, upgrades, services, outages, maintenance, repairs*, *sales*, *returns*, *cancellations*, etc., with some covering multiple issues on the same transcript. **Fig. 3** displays the distributions of the main issues amongst the 160K chat transcripts. The average and maximum lengths of the full chat and the constituent customer and agent transcripts were (314, 7295), (92, 4225) and (222, 4064) words respectively; while the average and maximum lengths of chat and the constituent customer and agent summaries were (75, 999), (34, 704) and (62, 576) words respectively. **Fig. 4** and **Fig. 5** show the histograms of lengths, i.e., the number of words contained in the text, for chat, customer and agent transcripts and summaries respectively in the 160K sample.

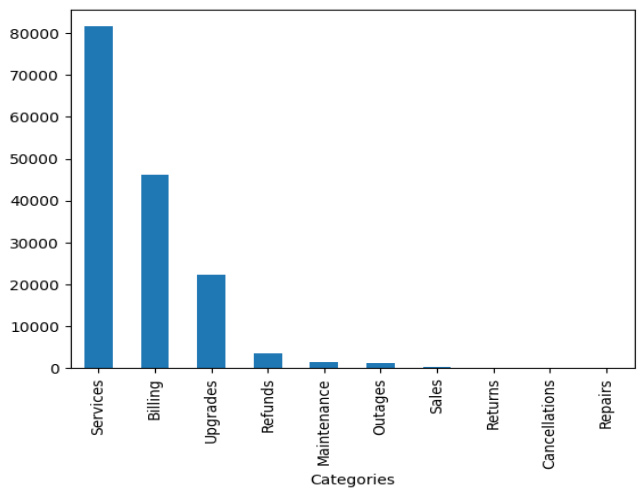

**Fig 3.** Distributions of Chat Transcripts' Main Issues/Topics

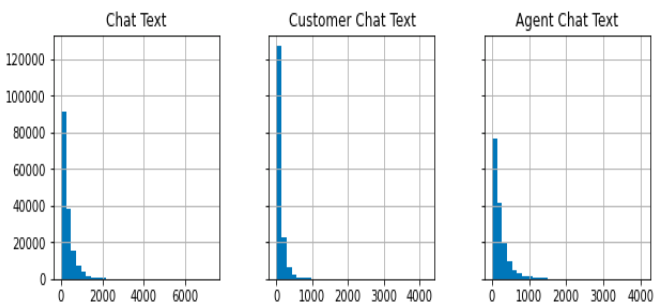

**Fig. 4.** Histograms of Chat, Customer and Agent Transcript word lengths in the 160K sample.

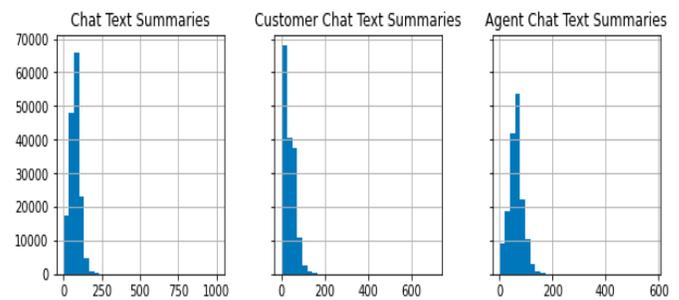

**Fig. 5.** Histograms of Chat, Customer and Agent Summary word lengths in the 160K sample.

In Phase II, the 160K transcripts (customer & agent) with their corresponding extractive summaries (from Phase I) were split into 3 sets, i.e., train, test and hold-out, with 150K, 5K and 5K samples respectively.

## C. Transformer/Language Models (LMs/LLMs)

In Phase I, we compared the performances of our extractive summarizer with those from another very popular, open-source extractive summarizer, namely, *Bert Extractive Summarizer* (BES) [45], using three pre-trained transformer Models: **BERT** (*bert-base-uncased*) [46], **GPT-2** (*gpt2-medium*) [47], and **XLNet** (*XLNet-base*) [48]. The *gpt2-medium* has 345 million parameters. **BERT** is a transformer encoder model and can be used as an extractive summarizer, while **GPT-2** and **XLNet** are decoder models that can abstract. We chose these 3 models as they can summarize well without any fine-tuning. Bert Extractive Summarizer generated summaries using the *period-restored* chat, customer, and agent transcripts from step 2 of the proposed procedure. Its *ratio* parameter was adjusted, using the number of words in the transcript, to ensure that its summaries were of comparable (shorter) lengths. The transcripts were summarized both *with (customer and agents)* and *without (full chat) channel separation*.

In Phase II, we fine-tuned pre-trained **T5** (*T5-small*) [50], **PEGASUS** (*Pegasus-large*) [51], **BART** (*Bart-large-xsum*) [52*]*, **Longformer2Roberta** (*longformer2roberta-cnn_dailymail-fp16*) [53] and **DialogLED** (*DialogLED-large-5120*) [54] models on train and test sets for abstractive summarization and validated their summaries on the hold-out set. The hyper-parameters of the models, e.g., *ecncoder_length, decoder_length, learning_rate, batch_size, num_beams, weight_decay, num_train_epochs, fp16, etc.,* were tuned to generate better summaries.

The open-source summarizers were used with their respective pre-trained models, each with its own *tokenizer, configuration, vocabulary,* and *checkpoints*.

## D. Evaluation Metrics

For automated evaluation, for measuring the *effectiveness* of our summarization and for comparing performances between extractive and abstractive summarizers, we used the metrics *BLEU* and *ROUGE-l scores*. We determined the *efficacy* of our *punctuation restoration* algorithm in Phase I using *accuracy_score* (sklearn.metrics).

The *efficiency* of a summarizer is important to real world applications. For Phase I, we measured the *efficiency* of our extractive summarizer by recording the time taken by each of the 10 steps of our proposed procedure. We also compared the *efficiency* of our *summary generation* through *sentence extraction algorithm* (step 7) with those of the open-source extractive summarizers by comparing the *total time* taken by each to *summarize* all of chat, customer & agent transcripts in the 160K sample. For Phase II, we compared the *efficiency* of the five abstractive summarizers by their *average fine-tuning times* for customer and agent transcripts.

### E. Results and Discussions: Automated Evaluation

We automatically evaluated the proposed hybrid strategy across all 3 phases. Here, we present our findings from the 3 phases with discussions.

*1) Phase I: Extractive Summarizers' Comparisons*

TABLE I shows results from Phase I and compares the *effectiveness* and *efficiency* of the indigenous (our) extractive summarizer for shorter summaries (~5 sentences) with those from the BERT Extractive Summarizer (BES) [12, 45] using three different pre-trained transformer models: **BERT** (*bert-base-uncased*), **GPT-2** (*gpt2-medium*), and **XLNet** (*XLNet-base*) over the 160K sample, on three different *evaluation metrics*. We compared all the extracted summaries with their corresponding *period-restored* original *transcripts* (from step 2 of our Phase I) for computing their *BLEU* and *ROUGE-l scores* as *we didn't have 160K manual summaries to compare them with*. Hence, the scores were low as the compared texts were of unequal lengths. However, the situation was the same for all the compared summarizers and the objective was to determine the extent of overlap between the extracted summaries from the different summarizers and the original transcripts. The *BLEU and ROUGE (ROUGE-l) scores*, for each type of transcript (chat/customer/agent), represent the *average* of the *BLEU* and *ROUGE* scores of all the summaries generated from the corresponding type of transcripts contained in the 160K sample. TABLE I shows that the indigenous extractive summarizer generated chat, customer and agent summaries with higher *average BLEU* and *average ROUGE scores*, *that is, overall better quality* than the summaries generated by BES using the three pre-trained models: **BERT**, **GPT-2**, and **XLNet** in approximately $\frac{1}{5}, \frac{1}{7}, \frac{1}{7}$ of the time taken by BES in summarizing the transcripts separately in the three cases. Thus, it establishes that our extractive summarizer is *more effective* (overlapped more) and *efficient* than BES for chat transcripts. *This justified the use of our extractive summaries for fine-tuning the 5 abstractive summarizers (transformer models) in Phase II*.

TABLE I also displays that the **BERT**-based BES performed better than the **GPT2-** and **XLNet**-based summarizers on the *average BLEU score, average ROUGE score* and the *summarization time* over the 160K chat, customer and agent transcripts, with the **XLNet**-based summarizer coming second.

In an **ablation study**, the *channel separation* component was removed from the indigenous extractive summarizer to understand how it affects performance. TABLE I illustrates that extractive summarization with *channel separation* generated more coherent summaries than without separation and the customer summaries were the most *effective*.

### TABLE I
### METRIC SCORES FOR EXTRACTIVE SUMMARIZERS

| Extractive Summarizer | Chat BLEU Score | Chat ROUGE Score | Customer BLEU Score | Customer ROUGE Score | Agent BLEU Score | Agent ROUGE Score | Total Summarization Time (secs) |
|---|---|---|---|---|---|---|---|
| **Indigenous Extractive Summarizer** | 0.20 | 0.52 | 0.30 | 0.63 | 0.24 | 0.55 | 17,334 (~5 hours) |
| BES(BERT) | 0.13 | 0.44 | 0.27 | 0.59 | 0.16 | 0.47 | 85,667 (~24 hours) |
| BES(GPT-2) | 0.12 | 0.40 | 0.26 | 0.57 | 0.15 | 0.44 | 124,161 (~35 hours) |
| BES(XLNet) | 0.12 | 0.41 | 0.26 | 0.58 | 0.15 | 0.45 | 118,199 (~33 hours) |

The *punctuation-restoration-accuracy* scores for chat, customer and agent summaries have also varied between 90 − 100% in all cases. The proposed summarizer is highly parameterized and provides more options than Bert Extractive Summarizer (BES).

*2) Phase II: Abstractive Summarizers' Comparisons*

TABLE II shows results from Phase II and compares the performances of the five abstractive summarizers: **T5** (*t5-small*), **PEGASUS** (*pegasus-large*), **BART** (*bart-large-xsum*), **Longformer2Roberta** (*longformer2roberta-cnn_dailymail-fp16)* and **DialogLED** (*DialogLED-large-5120*), on the hold-out set (5K). We compared all the customer and agent *abstracted* summaries with their corresponding *extracted summaries* from Phase I, for computing their *BLEU* and *ROUGE (ROUGE-l) scores*. The scores were higher as texts were of comparable lengths. TABLE II demonstrates that amongst the 5 abstractive summarizers, **PEGASUS** was most *effective* and *efficient*, i.e., generated customer and agent abstractive summaries *closest* to the extractive summaries with the highest *average BLEU & average ROUGE scores*, while taking the *least fine-tuning time*. **BART** came second. TABLE II further illustrates that **Longformer2Roberta** was more *effective* and *efficient* than **DialogLED** for our customer and agent transcripts. TABLE II also confirms that, *in our context*, **T5** was the least *effective* of all the fine-tuned summarizers for chat transcript summarization.

### TABLE II
### METRIC SCORES FOR ABSTRACTIVE SUMMARIZERS

| Abstractive Summarizer | Customer BLEU Score | Customer ROUGE Score | Agent BLEU Score | Agent ROUGE Score | Avg. Fine-tuning Time (secs) |
|---|---|---|---|---|---|
| T5 | 0.41 | 0.56 | 0.58 | 0.73 | 50,242 (~14 hours) |
| **PEGASUS** | 0.67 | 0.74 | 0.79 | 0.87 | 18,941 (~5.26 hours) |
| BART (XSum) | 0.62 | 0.72 | 0.67 | 0.83 | 19,214 (~5.34 hours) |
| Longformer2Roberta | 0.47 | 0.66 | 0.61 | 0.77 | 79,086 (~22 hours) |
| DialogLED | 0.46 | 0.63 | 0.55 | 0.74 | 114,574 (~32 hours) |

In an **ablation study**, the 5 above-mentioned transformer model based abstractive summarizers were used to summarize the chat transcripts in the hold-out set without *fine-tuning* any of the models, to measure the full impact of *fine tuning* for chat summarization in our context. TABLE III shows results from our ablation study and demonstrates the following: i) **PEGASUS** performed the best amongst the 5 *untuned* abstractive summarizers for all transcript types on all metric scores except on the *average BLEU score* for customer transcripts for which **DialogLED** showed the best results; ii) untuned **DialogLED** had higher *average BLEU scores* but lower *average ROUGE scores* than **Longformer2Roberta** for the two different transcript types; iii) untuned **BART** was the least effective for our customer and agent transcripts and this was because it generated much shorter summaries, when compared to our extractive summaries. Furthermore, comparing the metric scores in TABLE II and TABLE III, we can conclude that on an average, fine-tuning improved the performance of an abstractive summarizer on all transcripts by ~**8** times, on customer transcripts by ~**5** times and on agent transcripts by ~**11** times, and the improvement on *average BLUE scores* was ~**5** times more than that on the *average ROUGE scores*. **BART** showed the most improvement in fine-tuning on our chat transcripts.

TABLE III
METRIC SCORES IN ABLATION STUDY

| Abstractive Summarizer | Customer BLEU Score | Customer ROUGE Score | Agent BLEU Score | Agent ROUGE Score |
|---|---|---|---|---|
| **T5** | 0.07 | 0.28 | 0.05 | 0.21 |
| **PEGASUS** | 0.16 | 0.43 | 0.17 | 0.41 |
| **BART (XSum)** | 0.03 | 0.21 | 0.01 | 0.16 |
| **Longformer2Roberta** | 0.09 | 0.38 | 0.08 | 0.33 |
| **DialogLED** | 0.17 | 0.36 | 0.14 | 0.29 |

*3) Phase III: Validation and Optimization*

We validated our results from Phase II, i.e., our conclusions from TABLE II, through *bandit testing* using *contextual multi-armed bandits* (CMAB), while finding an optimal or near optimal solution for improving the overall quality of the 5K hold-out summaries, with respect to the *average BLEU and ROUGE (ROUGE-l) scores*. We used 6 features, namely *lengths*, *length fraction* (*length/total length*), *dominant topic categories*, *% dominant topic contributions, number of dominant topic keywords, number of document words* for both customer and agent transcripts as their contexts. We considered the five summarizers: **T5** [50], **PEGASUS** [51], **BART** (XSum) [52], **Longformer2Roberta** [53] and **DialogLED** [54] as the five arms of the 5-armed contextual bandit.

We applied multiple strategies (policies) for the *exploration vs exploitation tradeoff* for the 5-armed contextual bandit using the *Contextual Bandits* [55] package. This package provides APIs to implement multiple CMAB algorithms (strategies) including *Bootstrapped Upper Confidence Bound (BUCB)*, *Logistic Upper Confidence Bound (LUCB)*, *Bootstrapped Thomson Sampling*, *Epsilon Greedy*, *Adaptive Greedy*, *Explore First*, *Adaptive Active Greedy*, *and Softmax*. Details of all these algorithms can be found in [44]. We tried out all the algorithms available in this package on the customer and agent transcripts present in the hold-out set. We filtered out transcripts from the 5K hold-out set for which metric scores from all five summarizers were 0. For rewards, we used both *BLEU* and *ROUGE* (*ROUGE-l*) scores separately. Through *bandit testing*, using CMAB on the hold-out set, we converged smoothly, automatically, and progressively towards the best performing abstractive summarizer among the 5 fine-tuned transformer models, while optimizing (or nearly optimizing) the average metric scores of the abstractive summaries. **Fig. 6, Fig. 7,** and TABLE IV show the results from the applications of all the algorithms on the hold-out set.

**Fig. 6** shows the application of all the different algorithms (strategies) for CMAB on the *customer transcripts* from the hold-out set in optimizing the *average ROUGE score*. It shows that the *LUCB* strategy performed the best on the customer transcripts. In addition, TABLE IV (row 7, col. 3) shows that the winning strategy optimized the *average ROUGE (ROUGE-l) score* for all customer transcripts to 0.89 by applying **PEGASUS** 4120 times, **DialogLED** 24 times, **T5** 20 times, **Longformer2Roberta** 16 times, and **BART** 11 times on 4191 (~5K) hold-out transcripts.

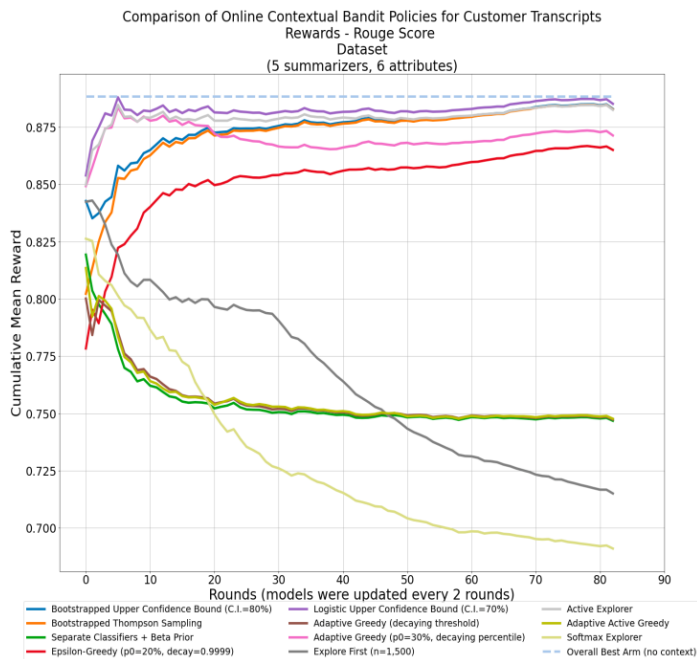

**Fig. 6.** Applications of CMAB Algorithms for Optimizing Customer Transcript Summarization and Identifying the Best Policy (Strategy).

**Fig. 7** shows the application of all the different algorithms for CMAB on the agent transcripts from the same hold-out set. It shows that the *LUCB* strategy also performed the best on the agent transcripts in optimizing the *average ROUGE (ROUGE-l) score*. Additionally, TABLE IV (row 7, col 5) shows that the winning strategy optimized the *average ROUGE score* for all agent transcripts to 0.87 by applying **PEGASUS** 4927 times, **DialogLED** 24 times, **T5** 20 times, **Longformer2Roberta** 16 times, and **BART** 11 times on 4998 (~5K) hold-out transcripts.

These experiments validate the results in TABLE II that **PEGASUS** is the *most effective* in summarizing the customer and agent chat transcripts as the winning strategies in both cases used **PEGASUS** [51] more frequently than any other fine-tuned abstractive summarizer in *maximizing* the average metric scores.

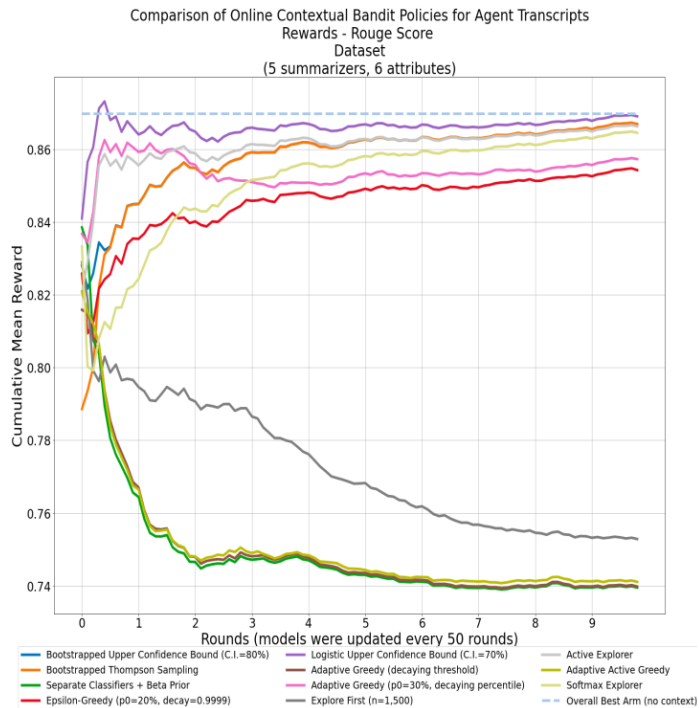

**Fig. 7.** Applications of CMAB Algorithms for Optimizing Agent Transcript Summarization and Identifying the Best Policy (Strategy).

TABLE IV
METRIC SCORES FROM BANDIT TESTING WITH CMAB

| Abstractive Summarizer | Customer BLEU Score | Customer ROUGE Score | Agent BLEU Score | Agent ROUGE Score |
|---|---|---|---|---|
| T5 | 0.41 | 0.56 | 0.58 | 0.73 |
| PEGASUS | 0.67 | 0.74 | 0.79 | 0.87 |
| BART (XSum) | 0.62 | 0.72 | 0.67 | 0.83 |
| Longformer2Roberta | 0.47 | 0.66 | 0.61 | 0.77 |
| DialogLED | 0.46 | 0.63 | 0.55 | 0.74 |
| CMAB [T5 + PEGASUS + BART + Longformer2Roberta (L2R) + DialogLED] | 0.67 (T5: 20 + PEGASUS: 4928 + BART: 11 + L2R: 16 + DialogLED: 24) | 0.89 (T5: 20 + PEGASUS: 4120 + BART: 11 + L2R: 16 + DialogLED: 24) | 0.79 (T5: 20 + PEGASUS: 4927 + BART: 11 + L2R: 16 + DialogLED: 24) | 0.87 (T5: 20 + PEGASUS: 4927 + BART: 11 + L2R: 16 + DialogLED: 24) |

TABLE IV extends TABLE II through reinforcement learning. It includes the results from bandit testing using CMAB (**Fig. 6** and **Fig. 7**). The last row in this table presents the *average BLEU* and *average ROUGE (ROUGE-l)* scores for customer and agent transcripts and the associated *components* (number of applications of the 5 different types of abstractive summarizers on the hold-out set) of the CMAB combination from the winning strategy in each case. It demonstrates that sometimes a strategy optimally combining multiple summarizers can be *better* or almost as *effective* as employing any single one. This can be easily explained by the fact that the most effective summarizer overall need not have the best performance (highest metric score) on each summarized transcript.

TABLE IV further corroborates that even if we *bypass* (or eliminate) the testing of the fine-tuned summarizers on the hold-out set in Phase II, we would still be able to identify the most *effective abstractive summarizer* for our chat transcripts through *bandit testing* and optimize or nearly optimize the overall quality of our summaries in production deployment.

Thus, **Fig. 6**, **Fig. 7** and TABLE IV, from Phase III, jointly lead to the following statement which is an important claim of our proposed strategy.

*Claim 6.1*: Bandit testing, utilizing the best strategy (policy) for Contextual Multi-armed Bandits, not only identifies the most effective LM/LLM-based fine-tuned abstractive summarizer for our chat transcript summarization, but also optimally improves the overall quality of summarization, in terms of average metric scores.

Based on these conclusions, a deployment and delivery framework consisting of either the un-optimized or the optimized CMAB strategies, i.e., policies (*LUCB*) for each of customer and agent transcripts, together with the 5 fine-tuned abstractive summarizers can be constructed for productionizing the automatic summarization of a large quantity of unknown, incoming chat transcripts on the fly in real time or in batch. If un-optimized policies are deployed, then they would learn to optimize in due course through *exploration and exploitation* just as was the case with the hold-out set in Phase III. However, the summaries would be evaluated against entire transcripts, in the absence of annotated summaries, as in Phase I. Conversely, if the optimized policies are instead deployed, based on the results from hold-out set in Phase III, then they would predict the next best summarizers for incoming customer and agent transcripts based on their contexts. **Fig. 8** shows this second case where an optimized CMAB strategy (policy) is used, with multiple transformer (LM/LLM) based fine-tuned abstractive summarizers, for chat transcript summarization in a production deployment environment.

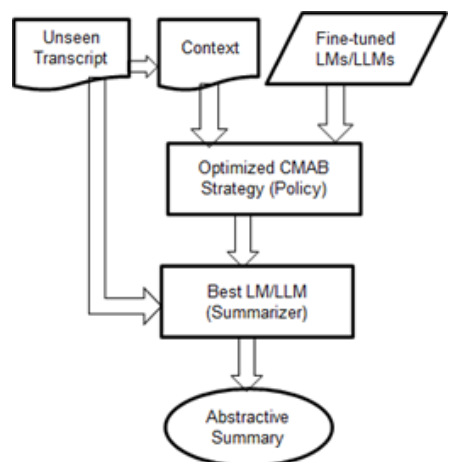

**Fig.8.** Production Deployment Framework

*4) Limitations*

There are two limitations associated with the proposed method, one related to its *evaluation procedure* and the other related to its *capability*. Limitation related to its automated evaluation originated from not having enough manually crafted reference, i.e., annotated summaries for the 160K chat transcripts under consideration. In the absence of a full set of reference summaries, we compared the extracted summaries with the *period-restored* and *longer* original transcripts (from step 2 of our extractive procedure) for computing their corresponding *ROUGE* and *BLEU scores*. So, the scores were slightly lower. However, this was done for the summaries from the proposed method as well as the three pre-trained language model driven Bert Extractive Summarizers to ensure consistency and similarity in the comparisons. Likewise, in the absence of manual reference (*annotated*) summaries, the abstractive summarizers were fine-tuned on extractive summaries, and we automatically compared the abstractive summaries with the extractive summaries using commonly used metric scores. However, this was done after verifying through both automatic and some manual evaluations that our extractive summaries were highly usable and readable. On the other hand, one limitation of its capability is that it *doesn't repair grammatical errors* (one of the six challenges associated with the chat transcripts), only reduces their numbers with fewer sentences, some post-processing and abstractive summarization through pre-trained transformer models (LMs/LLMs). This also explains the rationale behind the choice of the two *denoising* abstractive summarizers in Phase II for abstractive summarization.

Furthermore, it may be noted here that this research was started several years back, prior to the advent of the latest generation of instructional (prompt-based), all-purpose, transformer-decoder models, e.g., ChatGPT (GPT 3.5/4), Llama, etc. Consequently, for this version, we considered and tested slightly earlier generation of LMs/LLMs, i.e., non-instructional transformer-encoder-decoder models, which had been previously used in the literature for abstractive text (e.g., dialog) summarization (generation). However, the hybrid strategy is independent of the LMs/LLMs used in Phases II and III and is effective irrespective of which LMs/LLMs are used.

*F. Results and Discussion: Manual Evaluation*

The summaries generated by the proposed method are being manually verified and validated for content and readability by our business customers. The goal is to see if the summaries can be deemed generally useful for the purposes for which the transcripts were meant to be used in the specific use cases. The ongoing process is informal in nature and the evaluation is subjective. Feedback includes the following:

- If the chat was about one problem, then ~80% of the transcripts were capably summarized.
- Punctuations greatly improved the readability of the generated summaries.
- For *partially punctuated* extractive summaries, ~90 % of the periods, and for *fully punctuated* summaries, ~80% of all the punctuations were restored correctly.
- For the ~50 or so chat transcripts, where our extractive summaries were compared with manual summaries, the matches were found to be satisfactory.
- Our extractive summaries were more meaningful and readable than the summaries generated by their existing methods, namely *genism summarizer*, *pytextrank*, *pysummarization auto-abstractor* for their use cases.
- The abstractive summaries were readable, generally comparable to the extractive summaries and mostly expressed the main information content of the original transcripts.
- The abstractive summaries, due to fine-tuning, mostly maintained the same pronouns and verb forms (same person narratives) as the extractive summaries.
- The abstractive summaries didn't introduce opinions, interpretations, or subjective bias outside of those expressed in the extractive summaries.
- The abstractive summaries generated by **PEGASUS** and **BART** matched the extractive summaries more than **T5**, **Longformer2Roberta** and **DialogLED**.
- For the churn classification use case, our summaries captured the negative snippets adequately and helped validate the *churners*, that is, the results of churn classification.

### VII. CONCLUSION

In this paper, we have presented a hybrid summarization technique to address some of the challenges associated with domain-specific chat transcript summarization. We have combined *channel separation*, *topic modeling*, *sentence selection*, and *punctuation restoration* in *extractive summarization* with *fine-tuning* and *transfer learning for abstractive summarization* as well as *reinforcement learning* for *validation* and *optimization* to generate coherent and readable chat transcript summaries, to provide a better understanding of the customer complaints and the agent resolutions. The proposed summarizer is the *only* hybrid one that restores *full punctuation* to the summaries generated from either ill-punctuated or unpunctuated original chat transcripts. Finally, we have established the efficacy of the hybrid strategy through extensive experimentations and performance comparisons. We have validated some of our results through *bandit testing* using *contextual multi-armed bandits* based reinforcement learning. The hybrid method is very useful for large-scale deployment of chat transcript summarization when there is a dearth of *manually created reference,* i.e., *annotated summaries* for fine-tuning the abstractive summarizers and when the time spent on the summarization process is of essence.

### CONFLICT OF INTEREST STATEMENT

The author states that he is an employee of *Verizon* and this paper addresses work performed during the author's employment.